# Attention Deep Model with Multi-Scale Deep Supervision for Person Re-Identification

Di Wu, Chao Wang*, Yong Wu, Qi-Cong Wang* and De-Shuang Huang*, *Senior Member, IEEE*

*Abstract*—As an important part of intelligent surveillance systems, person re-identification (PReID) has drawn wide attention of the public in recent years. Many recent deep learning-based PReID methods have used attention or multi-scale feature learning modules to enhance the discrimination of the learned deep features. However, the attention mechanisms may lose some important feature information. Moreover, the multi-scale models usually embed the multi-scale feature learning module into the backbone network, which increases the complexity of testing network. To address the two issues, we propose a multi-scale deep supervision with attention feature learning deep model for PReID. Specifically, we introduce a reverse attention module to remedy the feature information losing issue caused by the attention module, and a multi-scale feature learning layer with deep supervision to train the network. The proposed modules are only used at the training phase and discarded during the test phase. Experiments on Market-1501, DukeMTMC-reID, CUHK03 and MSMT17 datasets. demonstrate that our model notably beats other competitive state-of-the-art models.

## I. INTRODUCTION

PERSON re-identification (PReID), which purpose is to re-identify a specific pedestrian of interest taken by multiple cameras or a single camera across different times in a camera network, has been extensively studied in recent years. Due to its significance in the application of intelligent video surveillance and security system, the task has attracted considerable interests in the computer vision community. However, the task is still difficult thanks to the large variations on captured pedestrians such as clothes, pose, illumination and uncontrolled complex background. To improve the PReID performance, extensive research has been reported in recent years. These works can be roughly divided into two types: a) exploiting discriminative features to represent the pedestrians' appearance, and b) learning a suitable distance metric for better computing the similarities between the paired person images. In the first category, the classical discriminative descriptors include the LOMO [1], ELF [2] and LBP [3]. In the second category, supervised learnings with the labeled images are used to obtain distance metrics function, such as LMNN [4], ADMM [5], LFDA [6] and PCCA [7]. However, these methods extract features and learn distance metrics separately, without considering their capabilities into one uniform structure.

Recently, with the successful development of deep learning technology, many deep learning approaches take advantage of Convolutional Neural Network (CNN) to learn discriminative and robust deep features for PReID. These approaches usually include two key components, *i.e.*, deep architecture and the objective functions. Early deep networks include VGGNet [8], ResNet [9] and DensNet [10]. Most recently, attention mechanism is favored by researchers in deep learning domain, to just a few, such as SENet [11], CBAM [12], BAM [13] and SKNet [14]. These models introduce the attention module into the state-of-the-art deep architectures to learn the spatial information and relationship between channels. In general, the softmax scores produced by the attention module are multiplied by the original features to output the final emphasized features. As a part of whole features, the un-emphasized features are also important to enhance the discriminative ability of descriptors, especially when they contain the body information. We argue that the un-emphasized features should be treated as emphasized features to help to learn the final descriptors. However, few of current PReID works have considered this issue.

The idea of utilizing middle-level features of a deep framework has been investigated, and has proved to be useful for object detection [15] and segmentation [16]. The strategy is first used by [17] for PReID. They used the deep supervision operation to train the combined embeddings of multiple convolutional network layers. Experimental results show the

This work was supported by grants from the National Science Foundation of China, Nos. 61520106006, 61732012, 61861146002, 61772370, 61702371, 61672203, 61772357, 61672471, 61873246 and 61672382, China Postdoctoral Science Foundation Grant, Nos. 2017M611619, and supported by the "BAGUI Scholar" Program of Guangxi Province in China. (*Corresponding author: De-Shuang Huang, Qi-Cong Wang and Chao Wang*)

Di Wu, Chao Wang, Yong Wu and De-Shuang Huang are with Institute of Machine Learning and Systems Biology, School of Electronics and Information Engineering, Tongji University, Caoan Road 4800, Shanghai 201804, China. (e-mail: wudi_qingyuan@163.com; dshuang@tongji.edu.cn)

Qi-Cong Wang is with department of computer, xiamen university (e-mail: qcwang@xmu.edu.cn)

effectiveness of the strategy. Yet, they fused embeddings at the lower and higher layers for training and test, which reduces the efficiency of the framework.

Recent works [17-19] have shown that multi-scale features learning can help to enhance the robust property of descriptors. Chen et al. [18] introduced a deep pyramidal feature learning framework which contains *m* scale-specific branches for multi-scale deep feature learning. More specifically, each branch learns one scale in the pyramid. Besides, they used a scale-fusion branch to learn the complementary of combined multi-scale features. Qian et al. [19] explored different resolution levels of filters to learn pedestrian features at multiple locations and spatial scales. These methods have shown great benefits on performance in PReID. However, using multi-branches to obtain multi-scale deep features increases the complexity of the framework.

By reviewing the current PReID works, we can find that the capability of a deep architecture could be improved by introducing the following strategies: (1) attention mechanisms; (2) middle-level features for deep supervision; (3) multi-scale features learning. Nevertheless, using attention mechanisms may cause the loss of important feature information. In addition, introducing the middle-level features to the final descriptors for deep supervision and adding the multi-scale features learning lower the efficiency of the model. In the PReID domain, these issues are barely considered. Therefore, in this work, we propose an auxiliary training based deep multi-scale supervision attention model to deal with the above issues.

As shown in Figure 1, the backbone of the proposed model is ResNet-50, which is used to extract different hierarchical deep features from the input image. The whole model is trained by a ranked triplet loss and four classification losses. To tackle the feature information losing issue caused by the attention module, we introduce a reverse attention block into the middle-level features learning. The block outputs probability scores that are complementary to the softmax scores produced by attention block. We do the dot product between the probability scores and the original features, thus, the un-emphasized features become the emphasized features. Then these features are passed through the pooling and concatenated to perform the classification task together. To deal with the efficiency problem, we propose a multi-scale layer to learn multi-scale information before the deep supervision operation. The multi-scale layer consists of several lightweight single scale convolution kernels, which learns the multi-scale information from horizontal and vertical directions, respectively. The deep supervision operators with multi-scale information learning only assist the deep framework to learn in the training phase, which are discarded these operators in the test phase. In this way, the efficiency of the deep framework is improved.

The contributions of this work are summarized as followings:
1) We propose a reverse attention block to remedy the non-salient feature information loss issue caused by the attention block.
2) We propose a lightweight multi-scale features learning block to perform deep supervision, which helps to learn more discriminative deep feature with multi-scale information.
3) The proposed operators above are only used during the training phase and abandoned for testing, thus improving the effectiveness of the inference network.
4) The proposed architecture outperforms other most recent state-of-the-art models on the four PReID datasets, including Market-1501, DukeMTMC-reID, CUHK03 and MSMT17.

## II. RELATED WORKS

In this section, we review some related works in the deep PReID domain. Then we introduce some recent attention deep models related to our method. Finally, some multi-scale deep models for PReID are briefly described. For more information about deep learning-based methods for PReID, we refer the readers to read [20].

A large number of deep learning-based methods have been proposed for PReID. There mainly exist three types of deep PReID models, *i.e.* identification model [21] [22] [23], verification model [24-34] and distance metric learning model [35] [36] [37]. The identification model formulates the PReID task as a classification problem. A fusion feature network was proposed by Wu et al. [21]. Through the supervision of identification loss, the model uses hand-crafted features to constrain the deep descriptors in the backpropagation phase. The verification model takes a pair of images as input, which calculates a score to represent the similarity of the paired images. Li et al. [24] proposed a verification model named filter pairing neural network (FPNN) for PReID. The distance metric learning model makes the relative distance between positive pairs smaller than that of negative pairs. Alexander et al. [37] proposed a hard mining method within a mini-batch for triplet loss. Wang et al. [38] proposed the Ranked List Loss (RLL) for building a set-based similarity structure by exploiting all instances in the gallery of deep metric learning. In this work, we combine the identification and Ranked List Loss to supervise the carefully designed deep architecture.

Attention is an important part of one person's perception. Inspired by this, there have been several works that try to incorporate attention mechanisms to state-of-the-art deep classification models to improve the performance of them. Hu et al. [39] introduced a compact 'Squeeze-and-Excitation' module to exploit the relationships between inter-channels. In the compact module, the channel-wise attention was computed by the global average-pooling features. By introducing spatial attention, Woo et al. [12] further proposed a convolution block attention module to simultaneously exploit channel-wise and spatial attention. They first used max-pooling and average-pooling operations to aggregate the channel information. Then these pooling features were concatenated and convolved to generate a 2-dimensional spatial attention map. The attention models above have been proven to be efficient for improving the performance of the deep network. However, introducing the attention mechanism may lead to the feature information loss problem. In our model, the attention module is also composed of the channel and spatial attention. Moreover, we introduce the reverse attention module to remedy the feature information loss situation.

Some works proposed to use multi-scale leaning for PReID.

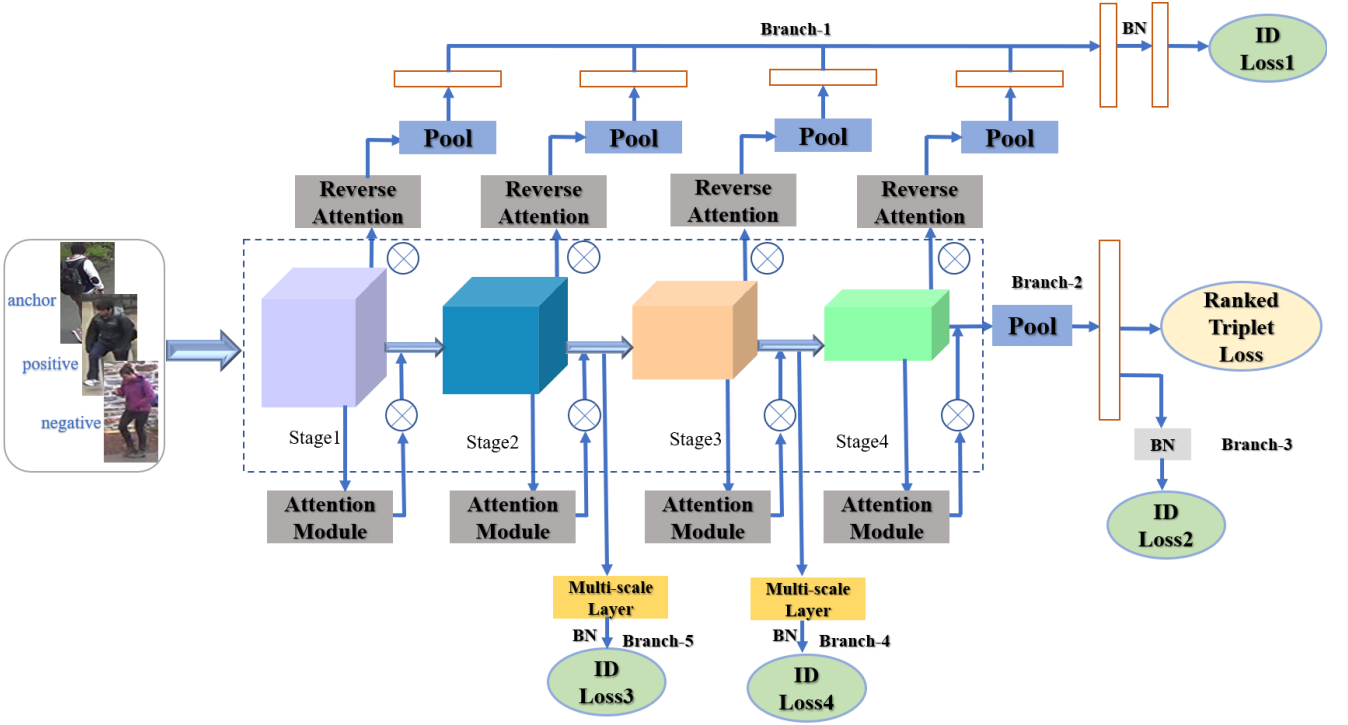

Fig. 1. Architecture of the proposed model: we use ResNet-50 as its backbone network. The architecture consists of five branches. Branch-1 with reverse attention block learns the feature information lost by attention block. By using the triplet and classification losses, branch-2 and branch-3 learn the global descriptors, respectively. Deep supervision with multi-scale feature learning is performed by branch-4 and branch-5. The model is supervised by four classification loss functions and one triplet loss function.

Liu et al. [40] proposed a multi-scale triplet deep architecture that learns deep features of a pedestrian at different scales. Specifically, the architecture integrated shadow and deep networks to produce low-level and high-level appearance features from images, respectively. Wang et al. [17] introduced a deeply supervised method for PReID. They used the pooling feature map of each stage to produce an embedding for each stage. Then these embeddings were fused by using a weighted sum. Both these methods show potential for PReID. Yet, since the multi-scale modules are inserted into deep architecture for training and inference, the whole network becomes computationally. Aiming at this problem, we propose a lightweight multi-scale feature learning module with deep supervision operations to help the deep model capture the multi-scale information.

### III. PROPOSED METHOD

In this section, we present the carefully designed deep construct by first presenting the training network, and then describing the loss functions. Finally, the inference network is described.

*A. Training Network*

The overview of the proposed deep architecture is shown in Fig 1. We use ResNet-50 as the backbone network, in which the last spatial down-sampling operation, the original global average pooling and fully connected layers are removed. We then append the average pooling layer and classification linear layer at the end of the backbone network. As shown in Fig 1, we utilize the feature maps produced by stage1, stage2, stage3 and stage4 of the ResNet-50 to generate attention and reverse attention masks. Besides, for reducing the GPU memory cost of the training network, we select the feature maps from stage2 and stage3 to perform multi-scale deep supervision operations. The detailed structure of the multi-scale layer is shown in Figure 2. The whole architecture is supervised by five losses, *i.e.* identity (ID) loss1, ID loss2, ID loss3, ID loss4 and triplet loss. Similar to the strong baseline work [41], ranked triplet loss and ID loss2 are used to learn global feature descriptors. ID loss1 is to supervise the reverse attention branch. We adopt the ID loss3 and ID loss4 to perform the multi-scale deep supervisions.

*B. Attention Block*

Enlighten by the works of [12] and [13], the attention block adopted in our architecture consists of channel attention and spatial attention. The channel attention outputs a set of weights for different channels while the spatial attention concentrates on the informative part.

**Channel Attention:** The channel attention block contains an average pooling layer and two linear layers. The feature maps $\mathbf{M}$ produced by the end of each stage are first passed through the average pooling operator as below:

$$\mathbf{M}_C = AvgPool(\mathbf{M}) \quad (1)$$

where $\mathbf{M} \in \mathbb{R}^{C \times W \times H}$, $\mathbf{M}_C \in \mathbb{R}^{C \times 1 \times 1}$.

Then the two linear layers with batch normalization are used for estimating attention across channels from $\mathbf{M}_C$. The size of output is set to $C/r$, where the parameter $r$ represents the reduction ratio. To restore the channel number, the output of the second linear layer is set to $C$. After the linear layers, a batch normalization layer is appended to adjust the scale of the spatial attention output. Thus, the channel attention $ATTc$ can be computed as:

$$ATT_C = BN(linear1(linear2(\mathbf{M}_C))) \quad (2)$$

where $linear1$, $linear2$ and $BN$ represent the first and second linear layer, and batch normalization, respectively.

**Spatial Attention:** Spatial attention block is used to emphasize features at different spatial positions. The block contains two reduction layers and two convolutions layers. Through the first reduction layer, the dimension of feature maps $\mathbf{M} \in \mathbb{R}^{C \times W \times H}$ is reduced to $\mathbf{M}_S \in \mathbb{R}^{C/r \times W \times H}$ with $1 \times 1$ kernel size. Then the $\mathbf{M}_S$ is passed through the two convolution layers with $3 \times 3$ kernel size. Finally, the number of feature maps are decreased to $\mathbb{R}^{1 \times W \times H}$ by using the second reduction layer with $1 \times 1$ convolution operation. Similar to the channel attention block, we apply the batch normalization operation at the end of the spatial attention block. The spatial attention $ATTs$ can be written as:

$$ATT_S = BN(Reduction2(Conv2(Conv1(\mathbf{M}_S)))) \quad (3)$$

where the $Conv1$ and $Conv2$ represent the two convolution layers, $Reduction2$ is the second reduction layer.

**Combine the two attentions:** We integrate the channel attention and the spatial attention into:

$$ATT = \sigma(ATT_C \times ATT_S) \quad (4)$$

where $ATT$ represents the whole attention block, and $\sigma$ is the *Sigmoid* function.

*C. Reverse Attention*

The attention block generates one probability mask to jointly emphasize the channels, and features in different spatial locations. However, through this operation, some channels and features are suppressed. We argue that these suppressed channels and features may be informative, which helps to enhance the discriminative ability of the final descriptors. Based on this assumption, we introduce reverse attention to complement the attention block. The output of the reverse attention in our model can be expressed as:

$$ATT_R = 1 - \sigma(ATT_C \times ATT_S) \quad (5)$$

We use the feature maps generated by each stage to do dot product with $ATT_R$. In this way, these suppressed channels and features become emphatic. Then these emphasized feature maps at each stage are passed through a pooling layer and concatenated to perform the classification task.

*D. Deep Supervision with Multi-Scale Learning*

From Fig 1, we use the feature maps at stage 2 and stage 3 to perform deep supervision operations, respectively. The deep supervision helps learn the discriminative descriptors. Note that we add the deep supervision operations behind the attention block, which is conducive to get more accurate attention maps. Besides, we introduce a multi-layer to acquire the multi-scale information for deep supervision. The structure of the multi-layer is shown in Figure 2, we first partition the channels into equal four groups. Then the four groups are passed through four convolution operations with kernel sizes of $1 \times 3, 3 \times 1, 1 \times 5$ and $5 \times 1$, respectively. After the convolution operations, the four groups are concatenated into one group. The reasons why we choose one-dimensional convolution are as follows:

a) One dimensional convolution has less parameter and reduces GPU memory consumption.

b) One dimensional convolution operation can learn the pedestrian features from horizontal and vertical directions, respectively, which adapts the human visual perception.

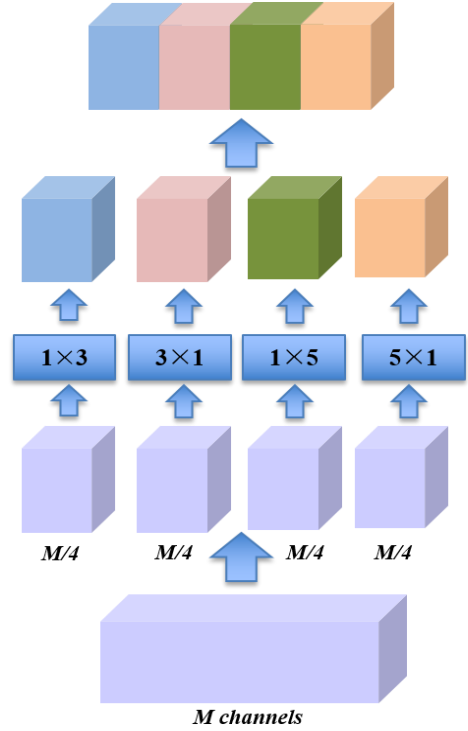

Fig. 2. The architecture of the proposed multi-scale layer.

*E. Loss Functions*

There have five loss functions in our model, *i.e.* four ID losses and one triplet loss. We employ the ranked list loss (RLL) proposed by [42] for metric learning and smoothing cross-entropy loss proposed by [43] for classification. The overall loss is the sum of them.

**Ranked List Loss:** We use the RLL to supervise the branch-2. The goal of RLL is to make the distance between negative samples larger than a threshold α, while the distance between positive samples closer than a threshold $α − m$, where m is a margin. The function can be written as:

$$L_m(x_i, x_j; f) = (1 - y_{ij})[\alpha - d_{ij}]_+ + y_{ij}[d_{ij} - (\alpha - m)]_+ \quad (6)$$

where $y_{ij} = 1$ denotes $x_i$ and $x_j$ are the same person, $y_{ij} = 0$, otherwise. $d_{ij}$ represents the Euclidean distance function, which is formulated as $d_{ij} = \|f(x_i) - f(x_j)\|$, $f(x_i)$ is the embedding function.

The nontrivial positive sample set can be displayed as:

$$P_{c,i}^* = \{x_j^c | j \neq i, d_{ij} > \alpha - m\} \quad (7)$$

The nontrivial negative sample set can be written as:

$$N_{c,i}^* = \{x_j^k | k \neq c, d_{ij} < \alpha\} \quad (8)$$

To pull all the nontrivial positive samples closer, we should minimize the following objective function:

$$L_P(X_i^c; f) = \sum_{X_i^c \in |P_{c,i}^*|} \frac{w_{ij}}{\sum_{X_i^c \in |P_{c,i}^*|} w_{ij}} L_m(X_i^c, X_j^c; f) \quad (9)$$

where $X_i^c$ represents a query image. $w_{ij}$ denotes the weighting positive pairs. It can be written as:

$$w_{ij} = \exp(T_n \times (\alpha - d_{ij})), X_j^k \in N_{c,i}^* \quad (10)$$

in which $T_n$ is a temperature parameter.

In order to push the distance between nontrivial negative samples larger than the threshold α, the following objective function should be minimized:

$$L_N(X_i^c; f) = \sum_{x_j^k \in |N_{c,i}^*|} \frac{w_{ij}}{\sum_{x_j^k \in |N_{c,i}^*|} w_{ij}} L_m(X_i^c, X_j^c; f) \quad (11)$$

where $w_{ij}$ is the weight of nontrivial negative samples. It can be represented as:

$$w_{ij} = \exp(T_n \times (\alpha - d_{ij})), X_j^k \in N_{c,i}^* \quad (12)$$

where $T_n$ is a temperature parameter.

Then the function of RLL can be written as:

$$L_{RLL}(x_i^c; f) = L_P(x_i^c; f) + \lambda L_N(x_i^c; f) \quad (13)$$

in which the $\lambda$ is the balance parameter, and we set it to 1.

**Smoothing Cross-Entropy Loss:** PReID can be regarded as a task of one-shot learning since the IDs in the test set are not included in the training set. Thus, preventing the overfitting issue for PReID models is very important. Label smoothing is an efficient method to alleviate the overfitting problem in the classification task, and has been effectively used in PReID domain. In this paper, we use label smoothing with cross-entropy loss to supervise the branch-1, branch-3, branch-4 and branch-5. Label smoothing function can be defined as:

$$q_i = \begin{cases} 1 - \frac{(N-1)\varepsilon}{N} & if\ i = y \\ \frac{\varepsilon}{N} & otherwise \end{cases} \quad (14)$$

where $y$ is the true label, $i$ represents the predicted label, $N$ is the number of training samples, $\varepsilon$ is a constant and we set it to 0.1. Then the function of smoothing cross-entropy loss is formulated as:

$$L_{ID} = \sum_{i=1}^{N} -q_i \log(p_i) \quad (15)$$

where $p_i$ denotes as the prediction logits of class $i$.

The overall loss function of the architecture can be written as:

$$L = \lambda_1 L_{RLL} + \lambda_2 L_{ID1} + \lambda_3 L_{ID2} + \lambda_4 L_{ID3} + \lambda_5 L_{ID4} \quad (16)$$

where $L$ is the total loss, $\lambda_1, \lambda_2, \lambda_3, \lambda_4$ and $\lambda_5$ are the balance coefficients.

### F. Inference Network

Our inference network is quite efficient and simple. As shown in Fig 3, in the test phase, we discard the multi-scale deep supervision, the reverse attention and the triplet branches, *i.e.* branch-1, branch-2, branch-4 and branch-5, and only use branch-3 to predict.

## IV. EXPERIMENTS

### A. Datasets

We perform the experiments on three popular PReID datasets, *i.e.* Market-1501 [44], CUHK03 [45] and DukeMTMC-reID [46]. The brief introductions of the datasets are presented below:

*Market-1501 dataset*: It contains 32643 images with 1501 pedestrians captured by at least two cameras and at most six cameras from a supermarket. The training set and testing set contain 12936 images of 751 IDs and 19732 images of 750 IDs, respectively.

*CUHK03 dataset*: The dataset contains 14097 images with 1467 pedestrians. It provides two bounding boxes in detection settings. One manually annotated by the human and the other automatically annotated by a detector. We conduct the experiments both in the two settings. Similar to [47], we divide the dataset into a training set with 767 pedestrians and a test set with 700 pedestrians.

*DukeMTMC-reID dataset*: It consists of 36411 annotated boxes with 1812 pedestrians captured by eight cameras.

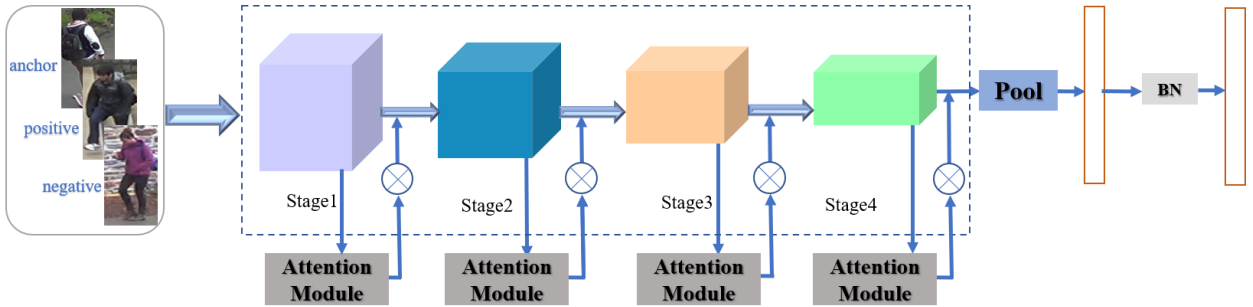

Fig. 3. The inference network.

Among the 1812 pedestrians, 1404 pedestrians appeared in more than two camera views, and the rest of pedestrians are treated as distractor identifications. The training set of this dataset consists of 16522 images of 702 pedestrians and the test set contains 17661 gallery images and 2228 query images.

*MSMT17 dataset*: The dataset was released in 2018, which was captured by fifteen cameras (three indoor cameras and twelve outdoor cameras) at different times. It consists of 4101 pedestrians with 126441 detected bounding boxes. The 1041 pedestrians with 32621 bounding boxes are used for training and the rest of 3060 pedestrians with 93820 bounding boxes are used for testing.

### B. Evaluation Metrics

Mean average precision (mAP) and cumulative match characteristic (CMC) are adopted as evaluation metrics to estimate the performance of our model. Besides, we report the Rank-1 and Rank-5 results. Both experiments of the datasets are conducted under single-query mode. Note that we don't use re-ranking in this work.

### C. Implementation Details

We use Pytorch to implement the proposed model. The ResNet-50 with pre-trained parameters on ImageNet is used as the backbone network. The hyperparameter reduction ratio $r$ of the attention block is set to 16. The commonly used data augmentation methods in PReID are followed. We resize all the images into 256×128. Then the size of outputs of stage1, stage2, stage3 and stage4 are 64×32, 32×16, 16×8 and 16×8, respectively. Random erasing [48] and horizontal random flipping are used as data augment methods for training. The batch size contains 64 images with 16 identities, in which each identity has 4 images. We train the architecture for 120 epochs. The $\lambda_1$, $\lambda_2$, $\lambda_3$, $\lambda_4$ and $\lambda_5$ described in Eq. (14) are set to 0.4, 0.1, 1, 0.03 and 0.03, respectively. We utilize adaptive moment estimation with an initial learning rate ($lr$) of $3.5 \times 10^{-5}$ to optimize the five loss functions. Following the work of [41], the $lr$ is then updated as the following rules:

$$lr(t) = \begin{cases} 3.5 \times 10^{-5} \times \frac{t}{10} & if\ t \leq 10 \\ 3.5 \times 10^{-4} & if\ 10 < t \leq 40 \\ 3.5 \times 10^{-5} & if\ 40 < t \leq 70 \\ 3.5 \times 10^{-6} & if\ 70 < t \leq 120 \end{cases} \quad (17)$$

The experiments are implemented on two TITAN XP GPUs.

### D. Comparison with state-of-the-art Methods

The proposed model is compared with the following methods, which includes PNGAN [49], PABR [50], PCB+RPP [51], SGGNN [52], MGN [53], G2G [54], SPReID [55], IANet [56], CASN [57], OSNet [58], BDB+Cut[59], P²-Net [60], and so on.

*Evaluation on market-1501 dataset.* The comparison results between our model and the state-of-the-arts on market-1501 dataset are shown in Table I. From the Table, we can observe that the proposed method outperforms the other competing models. Compared to Mancs which also utilizes attention and deep supervision operations, our model increases the mAP by + 6.7% and Rank-1 by +2.4%, respectively. The proposed model achieves mAP= 89.0%, Rank-1 accuracy= 95.5% and Rank-5 accuracy= 98.3% under single query mode, which validates the efficiency of it.

TABLE I
COMPARISON RESULTS ON MARKET-1501 DATASET.

| Method | Publication | Rank-1 | Rank-5 | mAP |
|---|---|---|---|---|
| PNGAN [49] | 2018ECCV | 89.4 | -- | 72.6 |
| PABR [50] | 2018ECCV | 90.2 | 96.1 | 76.0 |
| PCB+RPP [51] | 2018ECCV | 93.8 | 97.5 | 81.6 |
| SGGNN [52] | 2018ECCV | 92.3 | 96.1 | 82.8 |
| Mancs [61] | 2018ECCV | 93.1 | -- | 82.3 |
| MGN [53] | ACM MM18 | **95.7** | -- | 86.9 |
| FDGAN [62] | 2018NeurIPS | 90.5 | -- | 77.7 |
| DaRe [17] | 2018CVPR | 89.0 | -- | 76.0 |
| PSE [63] | 2018CVPR | 87.7 | 94.5 | 69.0 |
| G2G [54] | 2018CVPR | 92.7 | 96.9 | 82.5 |
| DeepCRF [64] | 2018CVPR | 93.5 | 97.7 | 81.6 |
| SPReID [55] | 2018CVPR | 92.5 | 97.2 | 81.3 |
| KPM [65] | 2018CVPR | 90.1 | 96.7 | 75.3 |
| AANet [66] | 2019CVPR | 93.9 | -- | 83.4 |
| CAMA [67] | 2019CVPR | 94.7 | 98.1 | 84.5 |
| IANet [56] | 2019CVPR | 94.4 | -- | 83.1 |
| DGNet [68] | 2019CVPR | 94.8 | -- | 86.0 |
| CASN [57] | 2019CVPR | 94.4 | -- | 82.8 |
| MMGA [69] | 2019CVPRW | 95.0 | -- | 87.2 |
| OSNet [58] | 2019ICCV | 94.8 | -- | 84.9 |
| Auto-ReID [70] | 2019ICCV | 94.5 | -- | 85.1 |
| BDB+Cut [59] | 2019ICCV | 95.3 | -- | 86.7 |
| MHN-6 [71] | 2019ICCV | 95.1 | 98.1 | 85.0 |
| P²-Net [60] | 2019ICCV | 95.2 | 98.2 | 85.6 |
| **Ours** | —— | **95.5** | **98.3** | **89.0** |

*Evaluation on CUHK03 dataset.* For the CUHK03, we adopt the protocol introduced by [47], in which 767 persons are utilized for training and the remain of 700 pedestrians are used for testing. The comparison results on CUHK03_detected and CUHK03_labeled settings are shown in Table II and Table III, respectively. We report mAP and Rank-1 accuracy under single query mode. From the two Tables, we find that the proposed method also beats all other compared state-of-the-art approaches, showing the efficiency of our method. Compared with Mancs, our method improves the mAP and Rank-1 accuracy by at least 13 percent, which further demonstrates the advantages of our model.

TABLE II
EXPERIMENT RESULTS ON CUHK03_DETECTED DATASET.

| Method | Publication | mAP | Rank-1 |
|---|---|---|---|
| MGN [53] | 2018ACMMM | 66.0 | 66.8 |
| PCB+RPP [51] | 2018ECCV | 57.5 | 63.7 |
| Mancs [61] | 2018ECCV | 60.5 | 65.5 |
| DaRe [17] | 2018CVPR | 59.0 | 63.3 |
| CAMA [67] | 2019CVPR | 64.2 | 66.6 |
| CASN [57] | 2019CVPR | 64.4 | 71.5 |
| OSNet [58] | 2019ICCV | 67.8 | 72.3 |

| Auto-ReID [70] | 2019ICCV | 69.3 | 73.3 |
| MHN-6 [71] | 2019ICCV | 65.4 | 71.7 |
| BDB+Cut [59] | 2019ICCV | 73.5 | 76.4 |
| P$^2$-Net [60] | 2019ICCV | 68.9 | 74.9 |
| **Ours** | —— | **75.3** | **78.8** |

TABLE III
EXPERIMENT RESULTS ON CUHK03_LABELED DATASET

| Method | Publication | mAP | R-1 |
| --- | --- | --- | --- |
| MGN [53] | 2018ACMMM | 67.4 | 68.0 |
| PCB+RPP [51] | 2018ECCV | -- | -- |
| Mancs [61] | 2018ECCV | 63.9 | 69.0 |
| DaRe [17] | 2018CVPR | 61.6 | 66.1 |
| CAMA [67] | 2019CVPR | 66.5 | 70.1 |
| CASN [57] | 2019CVPR | 68.0 | 73.7 |
| OSNet [58] | 2019ICCV | -- | -- |
| Auto-ReID [70] | 2019ICCV | 73.0 | 77.9 |
| BDB+Cut [59] | 2019ICCV | 76.7 | 79.4 |
| MHN-6 [71] | 2019ICCV | 72.4 | 77.2 |
| P$^2$-Net [60] | 2019ICCV | 73.6 | 78.3 |
| **Ours** | —— | **78.2** | **81.0** |

*Evaluation on the DukeMTMC-reID dataset.* As shown in Table IV, our proposed method achieves 79.2%/89.4% in mAP/Rank-1 on the DukeMTMC-reID dataset. Compared with the recent methods, we achieve the best results in mAP and Rank-1, which exceeds the state-of-the-art method MHN-6 by +2% and +0.3%, respectively.

TABLE IV
COMPARISON RESULTS ON DUKEMTMC-REID DATASET.

| Method | Publication | mAP | R-1 | R-5 | R-10 |
| --- | --- | --- | --- | --- | --- |
| G2G [54] | 2018CVPR | 80.7 | 88.5 | 90.8 | 66.4 |
| DeepCRF [64] | 2018CVPR | 84.9 | 92.3 | -- | 69.5 |
| SPReID [55] | 2018CVPR | 84.4 | 91.9 | 93.7 | 71.0 |
| PABR [50] | 2018ECCV | 82.1 | 90.2 | 92.7 | 64.2 |
| PCB+RPP [51] | 2018ECCV | 83.3 | 90.5 | 95.0 | 69.2 |
| SGGNN [52] | 2018ECCV | 81.1 | 88.4 | 91.2 | 68.2 |
| Mancs [61] | 2018ECCV | 84.9 | -- | -- | 71.8 |
| MGN [53] | 2018ACMMM | 88.7 | -- | -- | 78.4 |
| AANet [66] | 2019CVPR | 87.7 | -- | -- | 74.3 |
| CAMA [67] | 2019CVPR | 85.8 | | | 72.9 |
| IANet [56] | 2019CVPR | 87.1 | -- | -- | 73.4 |
| DGNet [68] | 2019CVPR | 86.6 | -- | -- | 74.8 |
| CASN [57] | 2019CVPR | 87.7 | -- | -- | 73.7 |
| OSNet [58] | 2019ICCV | 86.6 | -- | -- | 74.8 |
| Auto-ReID [70] | 2019ICCV | 88.5 | -- | -- | 75.1 |
| BDB+Cut [59] | 2019ICCV | 89.0 | -- | -- | 76.0 |
| P$^2$-Net [60] | 2019ICCV | 86.5 | 93.1 | 95.0 | 73.1 |
| MHN-6 [71] | 2019ICCV | 89.1 | 94.6 | **96.5** | 77.2 |
| **Ours** | —— | **89.4** | **94.7** | 96.0 | **79.2** |

TABLE V
COMPARISON RESULTS ON MSMT17

| Method | Publication | R-1 | mAP |
| --- | --- | --- | --- |
| BAT-net [72] | 2019ICCV | 79.5% | 56.8% |
| PCB [51] | 2018ECCV | 68.2% | 40.4% |
| IANet [56] | 2019CVPR | 75.5% | 46.8% |
| DG-Net [68] | 2019CVPR | 77.2% | 52.3% |
| OSNet [58] | 2019ICCV | 78.7% | 52.9% |
| Our | -- | **81.6%** | **59.4%** |

*Evaluation on the MSMT17 dataset.* MSMT17 is the largest dataset so far which is the most challenge due to its large-scale identities and distractors. To evaluate the performance of the proposed model, we further conduct the experiment on this dataset. As shown in Table V, our model achieves mAP=59.4% and Rank-1=81.6%, which also reaches the highest performance among the compared methods.

*E. Ablation Studies and Discussions*

We further implement some extra experiments to evaluate the effectiveness of each block of our model. All the ablation experiments are performed on CUHK03_labeled dataset under single query mode. The details of the experiment results of ablation studies are shown as below:

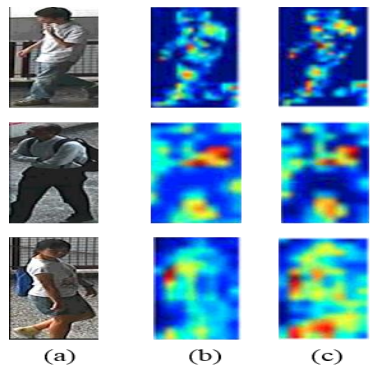

Fig. 4. Visualization of feature maps generated from the two setting. The column (a) represents the raw images, columns (b) and (c) represent the feature maps extracted without reverse attention and with reverse attention, respectively.

*Effectiveness of the reverse attention block.* In this setting, we discard the reverse attention and name the discarded model as Our/$_{reverse}$. The experiment results of the model on CUHK03_labeled dataset are shown in Table VI. From Table VI, we can observe that when discarding the reverse attention block, the performance of the model is decreased. More specifically, without reverse attention, the mAP and Rank-1 accuracy are reduced by -1.5% and -3.7%, respectively. To further prove the effectiveness of the reverse attention block, some feature maps generated by Our/$_{reverse}$ and Our models are presented in Fig 4. From Fig 4, we find that the fine details of feature maps are enriched when introducing the reverse attention block.

TABLE VI
THE EFFECTIVENESS OF ADVERSE ATTENTION BLOCK

| Model | mAP | R-1 | R-5 |
| --- | --- | --- | --- |
| Our/$_{reverse}$ | 76.7 | 77.3 | 91.3 |

| Our | 78.2 | 81.0 | 92.0 |

***Effectiveness of the multi-scale deep supervision block.*** To verify the effectiveness of the multi-scale deep supervision block, we use the model that discards branch-4 and branch-5 to implement the experiment. As shown in Table VII, when introducing the multi-scale deep supervision block, the performance of the discarded model Our/$_{supervision}$ is improved, increasing the mAP and Rank-1 accuracy by +1.3% and 1.9%, respectively.

TABLE VII
THE EFFECTIVENESS OF MULTI-SCALE DEEP SUPERVISION BLOCK

| Model | mAP | R-1 | R-5 |
|---|---|---|---|
| Our/$_{supervision}$ | 76.9 | 79.1 | 91.6 |
| Our | **78.2** | **81.0** | **92.0** |

## V. CONCLUSION AND FUTURE WORK

In this study, we introduce a multi-scale deep supervision with attention block deep model for person re-identification. We first design the reverse attention module to assist the attention module, then we introduce a multi-scale deep supervision block to learn features with multi-scale information as well as rectify the attention block. The experiment results on the three large datasets show that the proposed model exceeds the competitive methods. What's more, in this work, we only divided the channels into four groups in the multi-scale layers, we believe if we divide the channels into finer groups, the accuracy of PReID can be further improved.

Infrared-Visible PReID is a task of associating the same person across visible and thermal cameras. Most of the current studies tried to design the discriminative global features and ignored local and salient features for this issue. In the future, we would try to simultaneously use global and local features to address the cross-modality issue for PReID.

## REFERENCES

[1] R. Zhao, W. Ouyang, and X. Wang, "Learning Mid-level Filters for Person Re-identification," in *computer vision and pattern recognition*, 2014, pp. 144-151.
[2] D. Gray and H. Tao, "Viewpoint Invariant Pedestrian Recognition with an Ensemble of Localized Features," in *Computer Vision - ECCV 2008, 10th European Conference on Computer Vision, Marseille, France, October 12-18, 2008, Proceedings, Part I*, 2008.
[3] F. Xiong, M. Gou, O. Camps, and M. Sznaier, *Person Re-Identification Using Kernel-Based Metric Learning Methods*, 2014.
[4] M. Dikmen, E. Akbas, T. S. Huang, and N. Ahuja, "Pedestrian Recognition with a Learned Metric," in *Asian Conference on Computer Vision*, 2010, pp. 501-512.
[5] N. Martinel, A. Das, C. Micheloni, and A. K. Roychowdhury, "Temporal Model Adaptation for Person Re-Identification," *european conference on computer vision,* pp. 858-877, 2016.
[6] S. Pedagadi, J. Orwell, S. Velastin, and B. Boghossian, "Local Fisher Discriminant Analysis for Pedestrian Re-identification," in *IEEE Conference on Computer Vision and Pattern Recognition*, 2013, pp. 3318-3325.
[7] F. Jurie and A. Mignon, "PCCA: A new approach for distance learning from sparse pairwise constraints," in *IEEE Conference on Computer Vision and Pattern Recognition*, 2012, pp. 2666-2672.
[8] K. Simonyan and A. Zisserman, "Very Deep Convolutional Networks for Large-Scale Image Recognition," *Computer Science,* 2014.
[9] K. He, X. Zhang, S. Ren, and S. Jian, "Deep Residual Learning for Image Recognition," in *IEEE Conference on Computer Vision & Pattern Recognition*, 2016.
[10] G. Huang, Z. Liu, L. V. D. Maaten, and K. Q. Weinberger, "Densely Connected Convolutional Networks," in *CVPR*, 2017.
[11] J. Hu, L. Shen, and G. Sun, "Squeeze-and-Excitation Networks," *arXiv: Computer Vision and Pattern Recognition,* 2017.
[12] S. Woo, J. Park, J. Lee, and I. S. Kweon, "CBAM: Convolutional Block Attention Module," in *european conference on computer vision*, 2018, pp. 3-19.
[13] J. Park, S. Woo, J. Lee, and I. S. Kweon, "BAM: Bottleneck Attention Module," *arXiv: Computer Vision and Pattern Recognition,* 2018.
[14] X. Li, W. Wang, X. Hu, and J. Yang, "Selective Kernel Networks," in *computer vision and pattern recognition*, 2019, pp. 510-519.
[15] T.-Y. Lin, P. Dollar, R. Girshick, K. He, B. Hariharan, and S. Belongie, "Feature Pyramid Networks for Object Detection," in *2017 IEEE Conference on Computer Vision and Pattern Recognition (CVPR)*, 2017.
[16] J. Long, E. Shelhamer, and T. Darrell, "Fully Convolutional Networks for Semantic Segmentation," *IEEE Transactions on Pattern Analysis & Machine Intelligence,* vol. 39, pp. 640-651, 2014.
[17] Y. Wang, L. Wang, Y. You, X. Zou, V. S. Chen, S. Li*, et al.*, "Resource Aware Person Re-identification Across Multiple Resolutions," in *computer vision and pattern recognition*, 2018, pp. 8042-8051.
[18] Y. Chen, X. Zhu, S. Gong, and IEEE, "Person Re-Identification by Deep Learning Multi-Scale Representations," in *2017 IEEE International Conference on Computer Vision Workshop (ICCVW)*, 2017.
[19] X. Qian, Y. Fu, T. Xiang, Y. Jiang, and X. Xue, "Leader-based Multi-Scale Attention Deep Architecture for Person Re-identification," *IEEE Transactions on Pattern Analysis and Machine Intelligence,* pp. 1-1, 2019.
[20] D. Wu, S. Zheng, X. Zhang, C. Yuan, F. Cheng, Y. Zhao*, et al.*, "Deep learning-based methods for person re-identification: A comprehensive review," *Neurocomputing,* vol. 337, pp. 354-371, 2019.
[21] S. Wu, Y. C. Chen, X. Li, A. C. Wu, J. J. You, and W. S. Zheng, "An enhanced deep feature representation for person re-identification," in *Applications of Computer Vision*, 2016, pp. 1-8.
[22] T. Xiao, H. Li, W. Ouyang, and X. Wang, "Learning Deep Feature Representations with Domain Guided Dropout for Person Re-identification," in *Computer Vision and Pattern Recognition*, 2016, pp. 1249-1258.
[23] Y. Lin, L. Zheng, Z. Zheng, Y. Wu, and Y. Yang, "Improving person re-identification by attribute and identity learning," *arXiv preprint arXiv:1703.07220,* 2017.
[24] W. Li, R. Zhao, T. Xiao, and X. Wang, "DeepReID: Deep Filter Pairing Neural Network for Person Re-identification," in *IEEE Conference on Computer Vision and Pattern Recognition*, 2014, pp. 152-159.
[25] E. Ahmed, M. Jones, and T. K. Marks, "An improved deep learning architecture for person re-identification," in *Computer Vision and Pattern Recognition*, 2015, pp. 3908-3916.
[26] L. Wu, C. Shen, and A. v. d. Hengel, "Personnet: Person re-identification with deep convolutional neural networks," *arXiv preprint arXiv:1601.07255,* 2016.
[27] C. Kang, X. Yu, S.-H. Wang, D. Guttery, H. Pandey, Y. Tian*, et al.*, "A heuristic neural network structure relying on fuzzy logic for images scoring," *IEEE Transactions on Fuzzy Systems,* 2020.
[28] S. Wang, J. Sun, I. Mehmood, C. Pan, Y. Chen, and Y. D. Zhang, "Cerebral micro‐bleeding identification based on a nine‐layer convolutional neural network with stochastic pooling," *Concurrency and Computation: Practice and Experience,* vol. 32, p. e5130, 2020.
[29] S.-H. Wang, K. Muhammad, J. Hong, A. K. Sangaiah, and Y.-D. Zhang, "Alcoholism identification via convolutional neural network based on parametric ReLU, dropout, and batch normalization," *Neural Computing and Applications,* vol. 32, pp. 665-680, 2020.
[30] S.-H. Wang, Y.-D. Zhang, M. Yang, B. Liu, J. Ramirez, and J. M. Gorriz, "Unilateral sensorineural hearing loss identification based

[30] on double-density dual-tree complex wavelet transform and multinomial logistic regression," *Integrated Computer-Aided Engineering,* vol. 26, pp. 411-426, 2019.

[31] D.-S. Huang and J.-X. Du, "A constructive hybrid structure optimization methodology for radial basis probabilistic neural networks," *IEEE Transactions on neural networks,* vol. 19, pp. 2099-2115, 2008.

[32] D.-s. Huang, "Radial basis probabilistic neural networks: Model and application," *International Journal of Pattern Recognition and Artificial Intelligence,* vol. 13, pp. 1083-1101, 1999.

[33] D.-S. Huang, "Systematic theory of neural networks for pattern recognition," *Publishing House of Electronic Industry of China, Beijing,* vol. 201, 1996.

[34] C. Yuan, Y. Wu, X. Qin, S. Qiao, Y. Pan, P. Huang*, et al.*, "An effective image classification method for shallow densely connected convolution networks through squeezing and splitting techniques," *Applied Intelligence,* vol. 49, pp. 3570-3586, 2019.

[35] A. Hermans, L. Beyer, and B. Leibe, "In Defense of the Triplet Loss for Person Re-Identification," *arXiv preprint arXiv:1703.07737,* 2017.

[36] D. Cheng, Y. Gong, S. Zhou, J. Wang, and N. Zheng, "Person Re-identification by Multi-Channel Parts-Based CNN with Improved Triplet Loss Function," in *Computer Vision and Pattern Recognition*, 2016, pp. 1335-1344.

[37] A. Hermans, L. Beyer, and B. Leibe, "In Defense of the Triplet Loss for Person Re-Identification," *arXiv preprint arXiv:1703.07737,* 2017.

[38] X. Wang, Y. Hua, E. Kodirov, G. Hu, R. Garnier, and N. A. Robertson, "Ranked List Loss for Deep Metric Learning," in *computer vision and pattern recognition*, 2019, pp. 5207-5216.

[39] S. Woo, J. Park, J. Lee, and I. S. Kweon, "Cbam: Convolutional Block Attention Module," in *european conference on computer vision*, 2018, pp. 3-19.

[40] J. Liu, Z. Zha, Q. Tian, D. Liu, T. Yao, Q. Ling*, et al.*, "Multi-Scale Triplet CNN for Person Re-Identification," in *acm multimedia*, 2016, pp. 192-196.

[41] H. Luo, Y. Gu, X. Liao, S. Lai, and W. Jiang, "Bag of Tricks and a Strong Baseline for Deep Person Re-Identification," in *computer vision and pattern recognition*, 2019, pp. 0-0.

[42] X. Wang, Y. Hua, E. Kodirov, G. Hu, R. Garnier, and N. A. Robertson, "Ranked List Loss for Deep Metric Learning," *arXiv: Computer Vision and Pattern Recognition,* 2019.

[43] C. Szegedy, V. Vanhoucke, S. Ioffe, J. Shlens, and Z. Wojna, "Rethinking the Inception Architecture for Computer Vision," in *computer vision and pattern recognition*, 2016, pp. 2818-2826.

[44] L. Zheng, L. Shen, L. Tian, S. Wang, J. Wang, and Q. Tian, "Scalable Person Re-identification: A Benchmark," in *international conference on computer vision*, 2015, pp. 1116-1124.

[45] W. Li, R. Zhao, T. Xiao, and X. Wang, "DeepReID: Deep Filter Pairing Neural Network for Person Re-identification," in *computer vision and pattern recognition*, 2014, pp. 152-159.

[46] Z. Zheng, L. Zheng, and Y. Yang, "Unlabeled Samples Generated by GAN Improve the Person Re-identification Baseline in Vitro," *international conference on computer vision,* pp. 3774-3782, 2017.

[47] Z. Zhong, L. Zheng, D. Cao, and S. Li, "Re-ranking Person Re-identification with k-Reciprocal Encoding," in *computer vision and pattern recognition*, 2017, pp. 3652-3661.

[48] Z. Zhong, L. Zheng, G. Kang, S. Li, and Y. Yang, "Random Erasing Data Augmentation," *arXiv: Computer Vision and Pattern Recognition,* 2017.

[49] X. Qian, Y. Fu, T. Xiang, W. Wang, J. Qiu, Y. Wu*, et al.*, "Pose-Normalized Image Generation for Person Re-identification," in *european conference on computer vision*, 2018, pp. 661-678.

[50] Y. Suh, J. Wang, S. Tang, T. Mei, and K. M. Lee, "Part-Aligned Bilinear Representations for Person Re-identification," in *european conference on computer vision*, 2018, pp. 418-437.

[51] Y. Sun, L. Zheng, Y. Yang, Q. Tian, and S. Wang, "Beyond Part Models: Person Retrieval with Refined Part Pooling (and a Strong Convolutional Baseline)," in *european conference on computer vision*, 2018, pp. 501-518.

[52] Y. Shen, H. Li, S. Yi, D. Chen, and X. Wang, "Person Re-identification with Deep Similarity-Guided Graph Neural Network," *arXiv: Computer Vision and Pattern Recognition,* 2018.

[53] G. Wang, Y. Yuan, X. Chen, J. Li, and X. Zhou, "Learning Discriminative Features with Multiple Granularities for Person Re-Identification," *acm multimedia,* pp. 274-282, 2018.

[54] Y. Shen, H. Li, T. Xiao, S. Yi, D. Chen, and X. Wang, "Deep Group-Shuffling Random Walk for Person Re-identification," in *computer vision and pattern recognition*, 2018, pp. 2265-2274.

[55] M. M. Kalayeh, E. Basaran, M. Gokmen, M. E. Kamasak, and M. Shah, "Human Semantic Parsing for Person Re-identification," in *computer vision and pattern recognition*, 2018, pp. 1062-1071.

[56] R. Hou, B. Ma, H. Chang, X. Gu, S. Shan, and X. Chen, "Interaction-and-Aggregation Network for Person Re-identification."

[57] M. Zheng, S. Karanam, Z. Wu, and R. J. Radke, "Re-Identification With Consistent Attentive Siamese Networks," in *computer vision and pattern recognition*, 2019, pp. 5735-5744.

[58] K. Zhou, Y. Yang, A. Cavallaro, and T. Xiang, "Omni-Scale Feature Learning for Person Re-Identification," *arXiv: Computer Vision and Pattern Recognition,* 2019.

[59] Z. Dai, M. Chen, X. Gu, S. Zhu, and P. Tan, "Batch DropBlock Network for Person Re-identification and Beyond," *arXiv: Computer Vision and Pattern Recognition,* 2018.

[60] J. Guo, Y. Yuan, L. Huang, C. Zhang, J. Yao, and K. Han, "Beyond Human Parts: Dual Part-Aligned Representations for Person Re-Identification," *arXiv: Computer Vision and Pattern Recognition,* 2019.

[61] C. Wang, Q. Zhang, C. Huang, W. Liu, and X. Wang, "Mancs: A Multi-task Attentional Network with Curriculum Sampling for Person Re-Identification," in *european conference on computer vision*, 2018, pp. 384-400.

[62] Y. Ge, Z. Li, H. Zhao, G. Yin, S. Yi, X. Wang*, et al.*, "FD-GAN: Pose-guided Feature Distilling GAN for Robust Person Re-identification," in *neural information processing systems*, 2018, pp. 1222-1233.

[63] M. S. Sarfraz, A. Schumann, A. Eberle, and R. Stiefelhagen, "A Pose-Sensitive Embedding for Person Re-Identification with Expanded Cross Neighborhood Re-Ranking," *arXiv: Computer Vision and Pattern Recognition,* 2017.

[64] D. Chen, D. Xu, H. Li, N. Sebe, and X. Wang, "Group Consistent Similarity Learning via Deep CRF for Person Re-identification," in *computer vision and pattern recognition*, 2018, pp. 8649-8658.

[65] Y. Shen, T. Xiao, H. Li, S. Yi, and X. Wang, "End-to-End Deep Kronecker-Product Matching for Person Re-identification," in *computer vision and pattern recognition*, 2018, pp. 6886-6895.

[66] C. Tay, S. Roy, and K. Yap, "AANet: Attribute Attention Network for Person Re-Identifications," in *computer vision and pattern recognition*, 2019, pp. 7134-7143.

[67] W. Yang, H. Huang, Z. Zhang, X. Chen, K. Huang, and S. Zhang, "Towards Rich Feature Discovery With Class Activation Maps Augmentation for Person Re-Identification," in *computer vision and pattern recognition*, 2019, pp. 1389-1398.

[68] Z. Zheng, X. Yang, Z. Yu, L. Zheng, Y. Yang, and J. Kautz, "Joint Discriminative and Generative Learning for Person Re-identification," *arXiv: Computer Vision and Pattern Recognition,* 2019.

[69] H. Cai, Z. Wang, and J. Cheng, "Multi-Scale Body-Part Mask Guided Attention for Person Re-Identification," in *computer vision and pattern recognition*, 2019, pp. 0-0.

[70] R. Quan, X. Dong, Y. Wu, L. Zhu, and Y. Yang, "Auto-ReID: Searching for a Part-aware ConvNet for Person Re-Identification," *arXiv: Computer Vision and Pattern Recognition,* 2019.

[71] B. Chen, W. Deng, and J. Hu, "Mixed High-Order Attention Network for Person Re-Identification," in *international conference on computer vision*, 2019, pp. 371-381.

[72] P. Fang, J. Zhou, S. K. Roy, L. Petersson, and M. Harandi, "Bilinear attention networks for person retrieval," in *Proceedings of the IEEE International Conference on Computer Vision*, 2019, pp. 8030-8039.